\documentclass[letterpaper, 10 pt, conference]{ieeeconf}  % Comment this line out if you need a4paper

\IEEEoverridecommandlockouts                              % This command is only needed if 
\usepackage{amsmath} % assumes amsmath package installed
\usepackage{amssymb}  % assumes amsmath package installed
\usepackage{todonotes}
\let\labelindent\relax % deactivate ieee labelindent that conflicts with enumitem
\usepackage{enumitem}
\usepackage{mathtools}
\usepackage{rotating}
\usepackage{multirow}
\usepackage{siunitx}
\usepackage{caption}
\usepackage{subcaption}
\usepackage{float}
\usepackage{xcolor}
\usepackage{hyperref}
\usepackage{pifont}
\usepackage{cite}
\usepackage{microtype}

\usepackage{booktabs} %nice table design

\usepackage{algorithmic} %to display algorithms
\usepackage{algorithm2e}
\graphicspath{{graphics/}}

\usepackage{tabto} % to use tabstops in lists

\hypersetup{
    colorlinks=true,
    linkcolor=black,
    filecolor=magenta,      
    urlcolor=black,
    pdftitle={LaneDetectionDatasetSurvey},
    pdfpagemode=UseNone,
    }

\title{\LARGE \bf
Datasets for Lane Detection in Autonomous Driving:\\ A Comprehensive Review
}

\author{Jörg Gamerdinger$^{1}$, Sven Teufel$^{1}$, and Oliver Bringmann$^{1}$
\thanks{$^{1}$University of T\"ubingen, Faculty of Science, Department of Computer Science, Embedded Systems Group 
\tt\small {\{joerg.gamerdinger, sven.teufel, oliver.bringmann\} @uni-tuebingen.de}}
}%

\begin{document}
\maketitle
\thispagestyle{empty}
\pagestyle{empty}

\begin{abstract}
Accurate lane detection is essential for automated driving, enabling safe and reliable vehicle navigation across a variety of road scenarios. Numerous datasets have been introduced to support the development and evaluation of lane detection algorithms, each differing in terms of the amount of data,  sensor types, annotation granularity, environmental conditions, and scenario diversity. This paper provides a comprehensive review of 20 publicly available lane detection datasets, systematically analyzing their characteristics, advantages, and limitations. We classify these datasets based on key performance indicators such as sensor resolution, annotation types and diversity of road and weather conditions using a novel multidimensional metric for dataset quality. By identifying existing challenges and research gaps, we highlight opportunities for future dataset improvements that can further drive innovation in robust lane detection. This review serves as a resource for researchers seeking appropriate datasets for robust lane detection and contributes to the broader goal of advancing autonomous driving.
\end{abstract}

%%%%%%%%%%%%%%%%%%%%%%%%%%%%%%%%%%%%%%%%%%%%%%%%%%%%%%%%%%%%%%%%%%%%%%%%%%%%%%%%
\section{INTRODUCTION}
\label{sec:intro}
Road traffic injuries are the leading cause of death for young people aged 5 to 29~\cite{Death}. Autonomous vehicles are a promising approach to reducing the number of road accidents. To achieve this, a comprehensive and correct perception of the environment is required to increase safety. An important task for autonomous vehicles is lane detection, which is crucial for safe trajectory and motion planning. This information can then be used to determine valid trajectories. In order to achieve a robust and safe lane detection system, appropriate datasets with a sufficient size, diversity of scenarios, lighting and environmental conditions are required. The available datasets have different key aspects, strengths, and weaknesses that need to be considered before incorporating them into the development of lane detection systems. Therefore, we conducted a comprehensive review of existing and publicly available lane detection datasets to support decision-making on appropriate datasets in the development of lane detection systems. 

Currently, existing reviews of lane detection datasets only consider a subset of the available datasets, or lack sufficient discussion of the datasets. The lane detection dataset review by Shirke and Udayakumar (2019)~\cite{shirke2019lane} lacks several datasets and also includes datasets that cannot be used for lane detection, lacks appropriate references, and does not provide any discussion of the strengths and weaknesses of the datasets. Maghsoumi et al. (2024)~\cite{maghsoumi2024lane} presented a work on lane detection and tracking datasets. They extensively present detailed information for 14 datasets; however, some datasets are missing, and some datasets are not matching scientific standards. Moreover, their statistics are sometimes inaccurate as properties such as line types are analyzed using neural networks. Various general surveys about lane detection exist. Zakaria et al. (2023)~\cite{zakaria2023lane} includes 135 publications in total; however, only discusses 6 highly influential datasets. The survey by Li (2023)~\cite{li2023lane} also presents only a small subset of 10 datasets. Hao et al. (2023)~\cite{hao2023review} neglects the topic of datasets. The survey about monocular lane detection by He et al. (2024)~\cite{he2024monocular} lists 16 datasets; however, it lacks a sufficient discussion, and some datasets are not publicly available. The surveys about lane detection of Bi et al. (2025)~\cite{bi2025lane} and Koue et al. (2026) ~\cite{kou2026lane} both feature a list of 14 and 25 datasets; however, do not provide more details, comparison or discussion and include various unavailable datasets.
\begin{figure}[t]
    \centering
    \includegraphics[width=0.7\linewidth, clip, trim= 1cm 0cm 1cm 0cm]{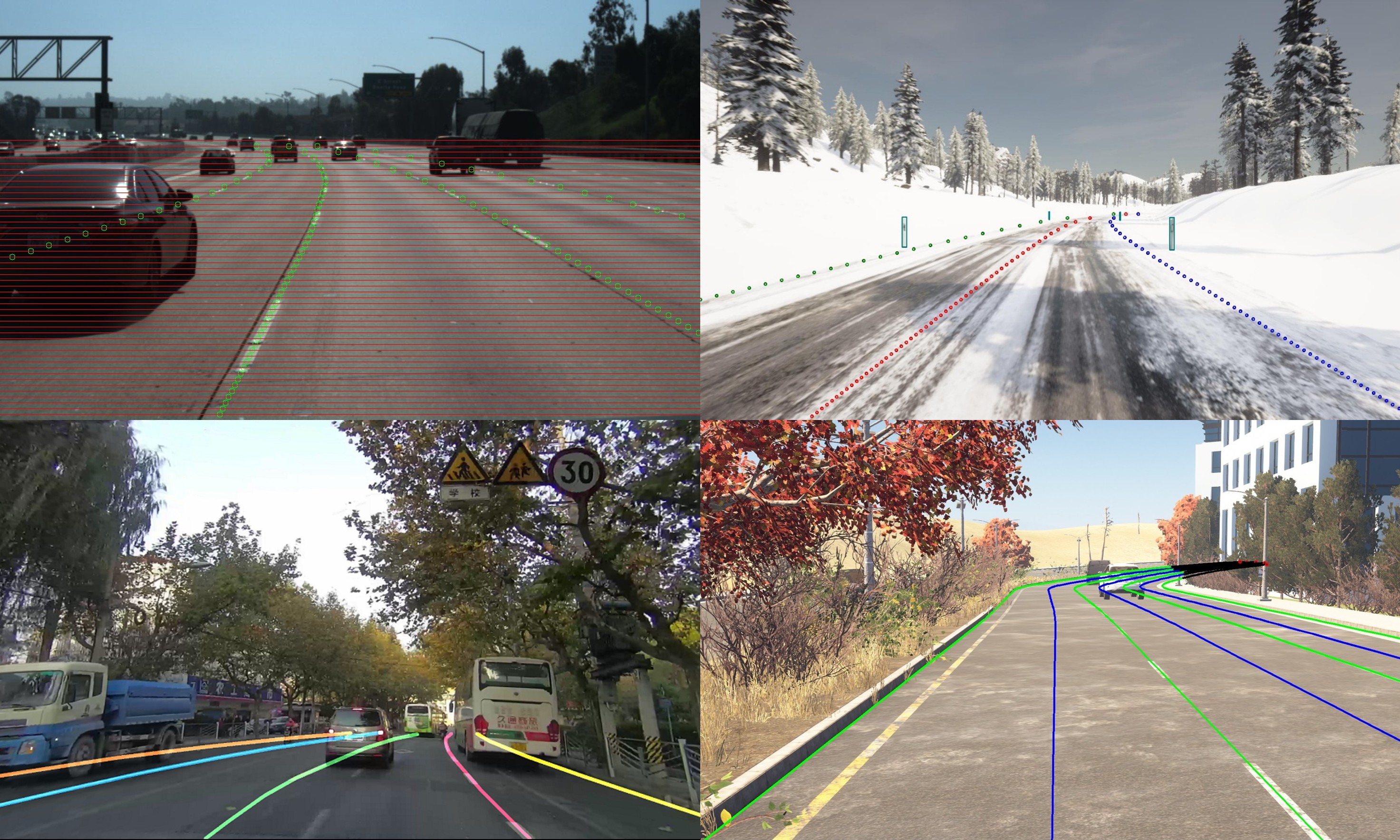}
    \caption{Exemplary annotated scenes from TuSimple~\cite{tusimple} (top left), CurveLanes~\cite{xu2020curvelane} (bottom left), SnowyLane~\cite{gamerdinger2025snowylane} (top right) and Gen-LaneNet~\cite{guo2020gen} (bottom right)} 
    \label{fig:example}
    \vspace*{-6mm}
\end{figure}
To overcome this lack of comprehensive reviews, we briefly review 20 lane detection datasets (see Sec.~\ref{sec:dataset_review}), propose an extensive evaluation strategy in Sec.~\ref{sec:eval} and discuss the results in Sec.~\ref{sec:results}. We then recommend which datasets are suitable for the development of robust lane detection systems. Lane detection can either be performed as a segmentation task to classify the drivable areas or as a detection task for lane markings which are then assigned to a specific lane. This review focuses on datasets that are publicly available and directly provide labels for training lane detection algorithms. Datasets that only provide semantic segmentation labels for drivable area are not in the scope of this review. Moreover, datasets that only provide map information which require a complex preprocessing before being suitable to train lane detection algorithms are only briefly presented.

\section{DATASET REVIEW}
\label{sec:dataset_review}

Within this section we provide details about published datasets for lane detection in chronological order based on their publication date, separated into real-world and synthetic datasets. An overview of all datasets with key information such as dataset size, sensors, and the environmental conditions is provided in Tab.~\ref{tab:overview}.

\subsection{Methodology}
For this review, we conducted an extensive literature research. We selected over 100 papers published since 2017 and indexed on Google Scholar. The search term ''lane detection'' was used to find papers about lane detection methods which use datasets, lane detection datasets that compare their dataset with other datasets and lane detection surveys, which contain sections on datasets. Using this research, we discovered over 40 datasets for the task of lane detection.

To provide meaningful statements about each dataset, only datasets that are publicly available are included in the evaluation process of this review. Hence, even Apolloscape~\cite{huang2018apolloscape} (unavailable since January 2025), is not included, as the dataset could not be analyzed. In addition, only datasets that are published with a corresponding paper at a conference or at least an arXiv preprint are considered, to avoid datasets that do not match scientific standards; hence, e.g., Jiqing expressway dataset~\cite{jiqing} is not included. One exception to this is the TuSimple dataset~\cite{tusimple}, which was part of the CVPR Autonomous Driving Challenge and hence is considered as scientific. Datasets such as OpenDenseLane~\cite{chen2022opendenselane}, which are not directly available outside of China, or datasets that do not provide labels, such as Yoo et al.~\cite{yoo2017robust}, are considered as not publicly available and hence are not part of this review. In addition, datasets only providing map information or semantic labels for road area are not part of this work; however, some highly influential are briefly presented at the end.

\subsection{real-world Datasets}

\subsubsection*{\textbf{ROMA}}
The ROMA dataset was published already in 2018 by Veit et al.~\cite{veit2008evaluation} and consists of 116 frames recorded in southern France. The recordings with a resolution of 1280$\times$1024 px. include labels as 2D segmentation masks on roads with only two lanes and two different types of markings. No adverse weather recordings are included.
\vspace*{2mm}
\subsubsection*{\textbf{KITTI Road}}
\label{subsec:kitti}
In 2013 Fritsch et al.~\cite{Fritsch2013ITSC} presented the KITTI Road dataset, which is a subset of 579 images from the KITTI dataset. The dataset includes all GPS information, images from stereo RGB cameras with a resolution of 1242$\times$375 px. and the LiDAR data. The dataset is divided into three categories (unmarked urban, two-way urban and multi-lane urban). The labels describe the road area as a 2D polygon. The images recorded on different days in an urban environment with clear weather contain a low traffic density, with most frames having no occlusion of road markings due to traffic participants. 
\vspace*{2mm}

\subsubsection*{\textbf{TRoM}}
\label{subsec:trom}
Liu et al.~\cite{liu2017benchmark} proposed the Tsingua Road Marking (TRoM) dataset in 2017, recorded in May 2016 in urban regions in Beijing, China. The dataset features 712 labeled frames with a resolution of 1280$\times$960 px. These frames include various weather conditions such as sunny, cloudy, varying rain intensities and fog and different times of day. The labels are provided as 2D segmentation masks of the road markings. Regarding lane markings, four different types are included.
\vspace*{2mm}

\subsubsection*{\textbf{TuSimple}}
\label{subsec:tusimple}
The TuSimple benchmark~\cite{tusimple} was introduced in 2017. The real-world dataset is recorded on US highways with 2-4 lanes, at different times of the day and in good weather. In addition to these variations, traffic density also varies between recordings. The recordings were made with a 1280$\times$ 720 px camera. A total of 7,000 clips of 20 frames each were recorded. For the benchmark, only 3,626 clips with one annotated frame per clip are provided for training. For testing, 2,728 clips with one frame per clip are used. The remaining 358 frames are used for validation. The TUSimple dataset uses 2D polylines as annotations. The dataset is widely used in research to compare lane detection approaches. 
\vspace*{2mm}

\subsubsection*{\textbf{CULane}}
\label{subsec:culane}
The CuLane dataset by Pan et al.~\cite{pan2018SCNN} was the first large-scale dataset for lane detection. The dataset consists of 133,235 annotated frames recorded in Beijing, China. The vehicles were equipped with a 1640$\times$590 px. RGB camera. The dataset contains different times of day, so that about \SI{20}{\percent} of the images are taken at night. In addition to the different times of day, the traffic density also varies between no vehicles and very busy scenes. Images are taken in urban areas of Beijing, but also in rural areas and on highways. 2D polylines are used for annotation and, unlike other datasets, occluded lane markings are also annotated, to allow for an estimation based on the visible parts.
\vspace*{2mm}%The size of the dataset makes it well suited for training lane detectors; however, although there are varying times of day, the dataset lacks adverse weather, which limits its applicability for robustness improvement.

% \newpage
\subsubsection*{\textbf{BDD100k}}
\label{subsec:bdd100k}
The BDD100k dataset by Yu et al.~\cite{yu2020bdd100k} was published in 2018. In total, the dataset consists of approximately 1.2 million images; however, as this dataset is applicable to multitask learning and object detection, not all images are labeled for the lane detection task. For lane detection, the dataset contains 100,000 labeled images. They are labeled as 2D polylines with additional information about the label type (dashed, double). In addition, the dataset contains labels for the detection of drivable areas. The data shows a wide variety of road types, as it includes urban, suburban and highway shots at different times of day and different weather conditions such as rain, snow and fog. 
\vspace*{2mm}

\subsubsection*{\textbf{PREVENTION}}
\label{subsec:prevention}
The PREVENTION dataset by Izquierdo et al. \cite{izquierdo2019prevention} was published in 2019. This dataset consists of 252,762 frames acquired with a 1920$\times$1080 px. RGB camera, a 32-layer LiDAR and three RADAR sensors. The vast majority of the shots were taken on highways, but there are also shots taken in urban environments with both in clear weather condition. The main task for this dataset is to predict intentions and trajectories in highway environments, but lane labels in the form of 3D polylines are also provided. 
\vspace*{2mm}

\subsubsection*{\textbf{Llamas}}
\label{subsec:llamas}
The Llamas dataset was published in 2019 by Behrendt and Soussan~\cite{llamas2019}. The dataset consists of 100,042 frames recorded on highways with up to 8 lanes and 6 classes, having a resolution of 1280$\times$717 px. with clear weather. The goal of the work was an unsupervised auto-labeling approach that shows remarkable results with annotations up to a range of over \SI{100}{\metre} for up to eight lanes. As the method is based on map information and LiDAR data, the dataset provides not only 2D polylines and segmentation masks for the images, but also 3D polylines. However, due to the auto-labeling various wrong annotations are included.
\vspace*{2mm}

\subsubsection*{\textbf{DET}}
\label{subsec:det}
The real-world DET dataset by Cheng et al.~\cite{det} was published in 2019, and is the only dataset using a dynamic vision sensor (DVS). The 5,424 raw DVS frames with a resolution of 1280$\times$800 px. were captured in Wuhan City, China under clear weather conditions. A total of 17,103 lanes are annotated using 2D segmentation masks with 1 to 4 lanes per frame and 5 marking types. The provided split is \SI{50}{\percent} for training, \SI{16.67}{\percent} for validation and \SI{33.33}{\percent} for testing. The dataset contains different traffic densities and scenarios such as bridges, tunnels and urban areas. 
\vspace*{2mm}

\subsubsection*{\textbf{TVTLane}}
\label{subsec:tvtlane}
The TVTLane dataset~\cite{zou2019tvt} by Zou et al. was presented in 2019. It is an extension of the TuSimple dataset and features a total of 39,460 labeled frames with a train/test split of \SI{78}{\percent}:\SI{22}{\percent}. Annotations are provided as 2D polyline. Besides the highway recordings from the TuSimple dataset, the TVTLane dataset also features recordings from urban and rural environments in China which are about \SI{25}{\percent} of the dataset. The additional recordings include 2-4 lanes and 4 lane marking classes. The images are downsampled to a joint resolution of 256$\times$128 px.
\vspace*{2mm}
%\newpage
\subsubsection*{\textbf{CurveLanes}}
\label{subsec:curvelanes}
The CurveLanes real-world dataset was published in 2020 by Xu et al.~\cite{xu2020curvelane}. As the name suggests, the dataset focuses on curvy road sections; thus, 135k of the total 150,000 images contain curved lanes. The images were taken with a 2650$\times$1440 px. RGB camera in urban and highway environments with 0 to 9 lanes. The dataset includes a variety of environmental conditions such as clear and cloudy days, wet roads, shaded areas, and nighttime recording but no direct adverse weather such as rain or fog. They provide a split of 100k images for training, 20k for validation and 30k for testing. 2D polylines are used for annotation.
\vspace*{2mm}

\subsubsection*{\textbf{VIL100}}
\label{subsec:vil100}
The VIL100 real-world dataset was published by Zhang et al.~\cite{zhang2021vil} in 2021. In total, the dataset consists of 100 sequences of 100 frames each, of which 97 are self-recorded using a monocular front-facing camera with 1920$\times$1080 px. or 1280$\times$720 px. The other 3 are taken from internet sources with a resolution of 640$\times$348 px. They use 10 classes of scenarios (e.g. crowded, curved road, fog, night, ...) to balance their data into an 80:20 train test split, and the recordings contain 1 to 5 lanes in 10 different lane types, annotated instance by instance with 2D polylines and lane type. In terms of variation, the dataset has different traffic densities, lighting conditions (day and night), and some data with fog.
\vspace*{2mm}

\subsubsection*{\textbf{Comma2k19LD}}
\label{subsec:comma2k19ld}
To allow a comprehensive evaluation with novel metrics, Sato et al.~\cite{sato2022towards} released the Comma2k19 LD dataset in 2022. To enable an application for improved metrics to assess safety or driving performance, the dataset includes not only 2D polyline lane annotations but also vehicle state information (position, orientation, speed). For this purpose, they used parts of the real-world Comma2k19 dataset~\cite{schafer2018commute} as a basis and annotated a total of 2,000 frames. The extracted parts are 100 frames from highways in California, United States, with a minimum speed of \SI{13.4}{\metre\per\second}. The image data has a resolution of 1920$\times$1080 px. and is recorded under clear weather conditions.
\vspace*{2mm}

\subsubsection*{\textbf{Once-3DLanes}}
\label{subsec:once3dlanes}
The Once-3DLanes by Yan et al.~\cite{yan2022once}, published in 2022, is based on the One Million Scenes real-world dataset, recorded in China. The recordings include highway, suburban and urban areas, as well as edge cases such as bridges and tunnels. The dataset consists of 213,111 images (202,111 for training, 3,000 for validation and 8,000 for testing) recorded at a resolution of 1920$\times$1080 px. The dataset contains different numbers of lanes ranging from 0 to 8 lanes. The annotations are provided as 3D polylines in camera coordinates. For further variation, the dataset includes day and night shots, as well as clear weather and different rain intensities. 
\vspace*{2mm}
\subsubsection*{\textbf{K-Lane}}
\label{subsec:klane}
The K-Lane dataset by Paek et al. \cite{paek2022klane} was published in 2022. Among most other lane detection datasets, K-Lane focuses on LiDAR-based lane detection. The dataset was acquired using an Ouster OS2 64-layer LiDAR sensor and a 1920$\times$1200 px RGB camera in Korea. In total, the dataset contains 15,382 labeled frames and is split into two subsets with 7,687 frames for training and 7,695 for testing. Both subsets contain recordings in different road conditions, such as day and night recordings, dense traffic with blocked lanes, as well as different lane types, but recordings in different weather conditions are not provided. The labels are provided as 3D polylines in the camera image. 
\vspace*{2mm}

\subsubsection*{\textbf{OpenLane}}
\label{subsec:openlane}
The OpenLane benchmark introduced by Chen et al.~\cite{chen2022persformer} in 2022 consists of 1000 segments with a total of 197,788 frames at a resolution of 1920$\times$1280 px. and 880k annotated lanes. The dataset is based on the real-world Waymo dataset~\cite{waymo}, which was recorded in Phoenix and San Francisco, United States. 
A unique feature of the dataset is that all lanes in a frame are annotated, resulting in up to 24 lanes per frame and 14 lane categories in total. They provide 2D and 3D polylines for annotation, as well as information on the type of lane marking, the position of the lane, and a unique ID per lane. The dataset provides a wide variety of scenarios, including day and night, dusk and dawn, highway, urban and suburban scenes. In addition, the dataset includes clear and cloudy weather, as well as rain and fog. Besides the lane annotation, the dataset features 2D bounding boxes for objects including a labeling how these vehicles are relevant for driving decisions. 
%\vspace*{2mm}

\newpage
\subsubsection*{\textbf{SDLane}}
\label{subsec:sdlane}
The SDLane dataset was released in 2022 by Jin et al.~\cite{jin2022eigenlanes}. They recorded a total of 43k frames with a resolution of 1920$\times$1208 px, divided into 39k for training and 4k for testing in South Korea. In terms of scenario variation, the dataset includes recordings from urban and highway sections, including tunnels and high density traffic with 0 to 5 lanes annotated using 2D polylines. The dataset features about \SI{75}{\percent} curved roads with up to \SI{89}{\degree}. The recordings are taken at daytime with clear weather.

\subsection{Synthetic Datasets}
\label{subsec:synthetic}

\subsubsection*{\textbf{Gen-LaneNet}}
\label{subsec:genlanenet}
The synthetic Gen-LaneNet dataset was proposed in 2020 by Guo et al.~\cite{guo2020gen}. The environment modeled in Unity is based on real-world maps in Silicon Valley, United States, and includes highways (6,000 images), urban areas (1,500 images), and residential areas (3,000 images) at three different times of day with clear weather. A single camera with a resolution of 1920$\times$1080 px. is used to capture the 10,500 images. In addition to the RGB image, a depth map, semantic segmentation, and 3D polylines are provided. The annotations are available up to a distance of \SI{200}{\metre} from the camera for 3 to 5 lanes.
\vspace*{3mm}

\subsubsection*{\textbf{SnowyLane}}
Gamerdinger et al.~\cite{gamerdinger2025snowylane} published the SnowyLane dataset in 2025. It is a synthetic dataset and is the first dataset which includes snow covering the road with three different intensities including snowfall and one subset without any snow. The road layout is based on rural roads in Southern Germany created as digital twin in CARLA~\cite{carla}. The dataset features a total of 80,000 frames with 1920$\times$1080 px. images and \SI{360}{\degree} LiDAR point clouds allowing for a multi-modal lane detection. The dataset also features road guidance posts and curvy roads. As the work focuses on rural roads only 2 lanes and 3 types of markings are included.
%\newpage
\subsection{Domain Adaptation}
The CARLANE dataset by Gebele et al.~\cite{NEURIPS2022_19a26064} was published in 2022 and is the only dataset that includes synthetic and real-world data to test domain adaptation. The dataset consists of three subsets with 168,000 unique frames from which 118,000 are labeled. The first subset (MoLane) includes a source synthetic dataset generated with CARLA~\cite{carla} and a target real-world dataset recorded with a model vehicle on an indoor test track. The synthetic dataset consists of 84,000 images. The model vehicle dataset consists of 46,843 frames. Images are captured using a 1280$\times$720 px. RGB camera. The CARLA dataset uses different randomizations of vehicle position, camera position and environment. The data is generated using five different maps from CARLA, including urban and highway areas. The synthetic dataset also includes different lighting conditions, road wetness, rain and fog. Annotations are provided as 2D polyline. The second dataset, called TuLane, combines 26,400 synthetic images with 6,408 images from the TuSimple dataset. The third subset, MuLane, combines 52,800 synthetic images with 12,536 images from the model vehicle and the TuSimple dataset. The number of lanes ranges between 1 and 4.
\vspace*{2mm}

\subsection{Map-based and Segmentation Datasets}
\label{subsec:maps}
real-world datasets such as Argoverse~\cite{chang2019argoverse} or nuScenes~\cite{caesar2020nuscenes} provide sensor data combined with HD map data. These datasets could be processed to be applied for lane detection, as lane annotations could be transformed from the map information into 2D image space. However, this requires the intrinsic and extrinsic camera matrix. In addition, deviations in the localization within the map lead to errors in the coordinate transformation from the HD map information to the image space. Therefore, real-world datasets with maps can be used for lane detection, but high-precision annotation cannot be guaranteed and additional effort is required to postprocess the dataset in order to obtain usable labels to train lane detectors. Depending on the method to determine the labels, deviations between works can occur. Hence, it is recommended to use datasets which are made for lane detection and provides labels that allow for comparability between different methods.

Semantic segmentation provides per-pixel class labels. This information can be used for lane detection if the class labels road and lane markings are available. However, highly influential semantic segmentation datasets such as Cityscapes~\cite{cityscapes} or Mapillary~\cite{neuhold2017mapillary} lack a class for lane markings and can be only used for drivable area detection. During the training process of lane detectors often binary segmentation masks for lane markings are calculated, which can be simplified using semantic labels. However, similar to the map-based data, preprocessing is required as for evaluation mostly 2D polylines or similar mathematical constructs are used which must be derived from the segmentation masks. This process can lead to inaccuracies and requires additional work which makes these datasets less suitable for lane detection. However, these datasets are well suited for drivable area segmentation or semantic understanding which are similar to lane detection but not in the scope of this work. 
%\vspace*{2mm}

\subsection{Non-Available Datasets}

\subsubsection*{\textbf{Caltech Lane}}
\label{subsec:caltech}
The Caltech Lane dataset proposed by Aly~\cite{aly2008real} in 2008 was the first public lane detection dataset. The dataset was collected in urban environments with straight and curved roads. In addition, the dataset includes areas with shadows on the road. In total, the dataset consists of four clips with a total of 1,224 frames with a resolution of 640$\times$480 px. and 4,172 labeled 2D lane boundaries. The dataset contains some data with road traffic, but only at a low density and no adverse weather.
\vspace*{2mm}

\subsubsection*{\textbf{SLD}}
\label{subsec:SLD}
In 2009 Borkar et al.~\cite{borkar2009robust} published a method for robust lane detection including the SLD dataset. The dataset features 2021 frames with a resolution of 720$\times$480 px. which were recorded in Atlanta, United States. Regarding the scenarios the dataset features data recorded on highways as well as urban environments with varying traffic densities and during day and nighttime. Unfortunately, the dataset is not publicly available; hence, the annotation type cannot be provided.
\vspace*{2mm}

\subsubsection*{\textbf{DIML}}
\label{subsec:diml}
In 2013, the real-world DIML dataset was published by Yoo et al.~\cite{yoo2013gradient}. The dataset contains 8,033 images with a resolution of 1280$\times$800 px., taken in South Korea. The dataset has a high scenario diversity as it includes images from highways, rural roads, and urban areas including tunnels. For all road types, DIML includes recordings at different times of day, including night, and environmental conditions are also present in the dataset, as about \SI{12}{\percent} of the frames are taken in rainy conditions. There is also a variation in traffic density. 
The dataset is not publicly available.
\vspace*{2mm}

\subsubsection*{\textbf{Yoo et al.}}
Yoo et al.~\cite{yoo2017robust} presented a small dataset recorded on highways in China in 2017. The dataset features 2,045 frames recorded with a monocular camera and sampled with a resolution of 320$\times$240 px. The dataset consists of five video streams in daytime, nighttime, tunnel, backlight and in a rainy day. These five subsets consists of highway recordings with 1 to 4 lanes and three different types of lane markings including white and yellow markings. Additionally, varying traffic densities are present but in general the traffic density is rather low.
\vspace*{2mm}

\subsubsection*{\textbf{VPGNet}}
\label{subsec:vpgnet}
The VPGNet dataset was published in 2017 by Lee et al.~\cite{Lee_2017_ICCV}. It was the first dataset to include adverse weather conditions. It was recorded in the urban environment of Seoul, South Korea, using a 1288$\times$728 px. camera. The dataset consists of 21,097 frames, divided into 14,783 frames for training and 6,314 frames for testing. The images contain different lane types with white, yellow, solid and dashed markings. The markers are annotated as 2D polygons. For scenario diversity they include day and night frames (about \SI{10}{\percent}). They also include data with clear weather, rain and heavy rain. The rain portion is about \SI{23}{\percent} of the data. 
\vspace*{2mm}

\subsubsection*{\textbf{ELAS}}
\label{subsec:elas}
The ELAS dataset by Berriel et al.~\cite{berriel2017ego} presented in 2017 is a real-world lane detection dataset recorded in three different cities in Brazil. Twenty sequences, including urban and highway areas with different lighting and weather conditions, were recorded for a total of 15k frames. They provide manually annotated 2D splines, but only for the ego lane. A GoPro Hero with 640$\times$480 px. was used for recording. Unfortunately, the dataset is not publicly available.
\vspace*{2mm}

\subsubsection*{\textbf{Bai et al.}}
Bai et al. \cite{bai2018deep} recorded a dataset for lane detection in 2018 to evaluate their multi-sensor lane detection framework. The dataset was recorded in North America and consists of two parts, a highway part with 22,073, 2,572, and 5,240 frames for training, validation, and testing, and an urban part with 16,918, 2,607, and 5,758 frames for training, validation, and testing. The dataset includes camera and LiDAR recordings, and the lane labels are given in both the 2D image space and BEV. As the dataset is not publicly available, no information about the sensor setup can be provided.
\vspace*{2mm}

\subsubsection*{\textbf{Apolloscape}}
Huang et al.~\cite{huang2018apolloscape} presented the apolloscape dataset in 2018 with recordings from Beijing, China. The recordings were taken using a test vehicle equipped with a stereo camera setup with 3384$\times$2710 px.
In total, the dataset for lane marking detection features 165,979 frames which are split into 132,189 training images and 33,790 testing images. For those images labels as 2D and 3D segmentation masks are provided. Regarding diversity, the dataset features varying traffic densities and 9 lane marking types including white and yellow lines. Moreoever, the dataset features different lighting conditions due to different times of day including nighttime recordings; however, adverse weather such as rain or fog are not part of the dataset.  

\vspace*{2mm}

\subsubsection*{\textbf{DSDLDE}}
\label{subsec:dsdlde}
In 2019 Lee and Moon~\cite{lee2018robust} proposed the DSDLDE dataset which consists of 48 video clips with a total of 33,323 frames recorded in the United States and South Korea. The images show a resolution of 1920$\times$1080 px. The dataset shows a high scenario diversity as it includes recordings from highways, urban areas and tunnels at day and nighttime with varying camera positions. Additionally, for day and nighttime clear weather, rain with varying intensities and recordings with snowfall are included. The labels are manually annotated as 2D polylines in the TuSimple format. The dataset is not publicly available.
\vspace*{2mm}

\subsubsection*{\textbf{Jiqing Expressway}}
\label{subsec:jiqing}
The Jiqing Expressway dataset was published in 2019 by Feng et al.~\cite{jiqing}. The images are 40 sequences taken on a single expressway in Jiqing, China, with a resolution of 1920$\times$1080 px. In total, the dataset consists of 210,610 images taken on both clear and cloudy days. In addition, some sections in tunnels are included to account for limited lighting conditions. The dataset contains 2 to 4 lanes annotated with 2D keypoints. 
\vspace*{2mm}

\subsubsection*{\textbf{Garnett et al.}}
To evaluate their 3D lane detection framework, Garnett et al. \cite{garnett20193d} generated a synthetic lane detection dataset using Blender in 2019. They used a randomized approach to generate different road scenes, which were then used to render 360$\times$480 px. RGB images. In total, 300k images were generated for training, 1k for validation and 5k for testing. The road labels are provided as 3D points in camera coordinates. Unfortunately, this dataset is not publicly available. In 2020, to study domain adaptation in lane detection, Garnett et al. \cite{garnett2020synthetic} generated a new dataset following the method in \cite{garnett20193d}. They generated 50k images using the camera parameters from different target domains such as TUSimple and Llamas. Unfortunately, this dataset is not publicly available either.
\vspace*{2mm}

\subsubsection*{\textbf{3DLanes}}
\label{subsec:3dlanes}
The 3DLanes dataset by Efrat et al.~\cite{efrat2020semi} is a real-world 3D lane detection dataset with a total of 327k frames from different geographic locations within an area of \SI{250}{\kilo\metre}. The dataset is split into 298k frames for training and the remaining frames are sub-sampled leading to a test set of 1,000 frames. For scenario diversity recordings were taken on highways and rural roads at different times of day under clear weather conditions. As the dataset is not publicly available, different information such as the camera resolution can not be provided.
\vspace*{2mm}

\subsubsection*{\textbf{TTLane}}
In 2020 Liang et al.~\cite{liang2020lane} proposed the TTLane dataset. The dataset consists of 13,200 frames with a resolution of 2058 $\pm$ 163 px. (width) $\times$ 1490 $\pm$ 215 px. (height) recorded in an urban environment. The varying resolution is due to the fact that the recordings were taken by about 200 different drivers with mobile phone cameras mounted on the vehicle dashboard. The annotations are provided as 2D Bézier curves, distinguished between occluded and visible lane markings. The dataset features different lighting conditions as well as sunny and rainy weather conditions. However, the dataset is not publicly available.
\vspace*{2mm}

\subsubsection*{\textbf{FusionLane}}
To evaluate their multi-modal lane detection framework called FusionLane, Yin et al. \cite{yin2020fusionlane} created a lane detection dataset based on the KITTI dataset \cite{kitti} in 2020. They labeled 437 camera images with a resolution of 1242$\times$375 px. were projected into BEV. In addition, also LiDAR projections in BEV are provided. Using rotations, they expanded their dataset to a size of 14,720 images. These images are recorded in urban environments with clear weather. Unfortunately, the dataset is not available.
\vspace*{2mm}

\subsubsection*{\textbf{OpenDenseLane}}
\label{subsec:opendenselane}
OpenDenseLane~\cite{chen2022opendenselane} was published in 2022 by Chen et al. This dataset is a road markings detection dataset that focuses on the detection of lanes, crosswalks, stop lines, curbs, and road signs such as directional arrows. OpenDenseLane was captured using two Ouster OS1 128-layer and one Hesai Pandar 128-layer LiDAR sensors and multiple cameras. In total, the dataset consists of 1,709 day and night driving scenarios on urban roads and highways, resulting in 57,227 frames. The labels are provided as 3D point labels for the points in the LiDAR point clouds. For the road class, the lane type is also labeled and for the road signs, which are mostly turn arrows, the arrow direction is labeled.
The labels are also provided as BEV image masks. Unfortunately, no information on the cameras used is provided and the camera images are not included in the dataset.
\vspace*{2mm}

\subsubsection*{\textbf{Simulanes}}
\label{subsec:simulanes}
Simulanes by Hu et al.~\cite{hu2022sim} is a synthetic dataset generated for domain adaptation using the CARLA simulator. The dataset has 16,344 frames with a resolution of 1280$\times$720 px. The resolution and annotation as 2D anchor points to fit a polyline correspond to the TuSimple dataset used as the target domain. The dataset has a range of scenarios as it is generated on 6 different CARLA maps including urban, suburban, rural and highway areas combined with different weather conditions such as clear weather, rain and fog. The dataset also includes a wide range of lanes (2 to 10 lanes) with 15 different classes of markings. The dataset provides a wide variety of adverse weather scenarios.  
\vspace*{2mm}

\subsubsection*{\textbf{LanEvil}}
Zhang et al. \cite{zhang2024lanevil} presented the LanEvil dataset in 2024. The main focus of this dataset is to evaluate the robustness of lane detection methods to common environmental lane image corruptions. Using the CARLA simulator, they generated 94 scenarios with 14 different types of corruption, such as road damage, reflections, shadows and obstacles. In total, the LanEvil dataset consists of 90,292 images with a resolution of 1280$\times$720 px. This dataset is divided into two subsets for training and testing, of 40,000 and 50,292 images respectively. The simulated scenarios show a high scenario diversity including urban roads and highways, as well as clear and rainy conditions.

\section{EVALUATION}
\label{sec:eval}

To enable a meaningful comparison of lane detection datasets, we introduce a multidimensional dataset quality score (DQS) $\in [0,1]$ for lane detection. Bouhsissin et al.~\cite{bouhsissin2023evaluating} demonstrated multi-criteria decision analysis to create a weighted score for ranking dataset in intelligent transportation using different key performance indicators (KPIs). Our score integrates five complementary KPIs similar to~\cite{bouhsissin2023evaluating}: dataset scale $S_{\text{scale}}$, diversity $S_{\text{div}}$, annotation richness $S_{\text{anno}}$, sensor richness $S_{\text{sensor}}$, and research impact $S_{\text{impact}}$. These scores are normalized and weighted to determine DQS. Our DQS framework allows for a variable weighting of these scores based on the primary goal that should be achieved using the dataset. E.g. if the model only should be applied to highway scenarios but requires a lot of data, the scale could be weighted higher than the diversity.

For the scale score $S_{\text{scale}}$ (see Eq.~\eqref{eq:scale}) a logarithmic normalization for the frames $F_d$ of the dataset is used to limit dominance of extremely large datasets.
\begin{equation}
S_{\text{scale},d} =
\frac{\log(F_d) - \min_d(\log(F_d)}
{\max_d(\log(F_d)) - \min_d(\log(F_d))}
\label{eq:scale}
\end{equation}
$S_{\text{div}}$ (see Eq.~\eqref{eq:s_div}) represents the diversity within the dataset including $E_d$ as number of environmental conditions, $A_d$ as number of areas used driving scenarios and $L_d$ the maximum number of visible lanes in dataset $d$ and an indicator $I_{\text{real}}$ for real-world data. 
The normalization constants are defined as $E_{\max}=7$, $S_{\max}=4$, and $L_{\max}=24$ representing the maximum numbers of the respective variables for normalization. 

\begin{figure*}
    \centering
    \includegraphics[width=.92\linewidth, trim = 2cm 1cm 2cm 1cm ]{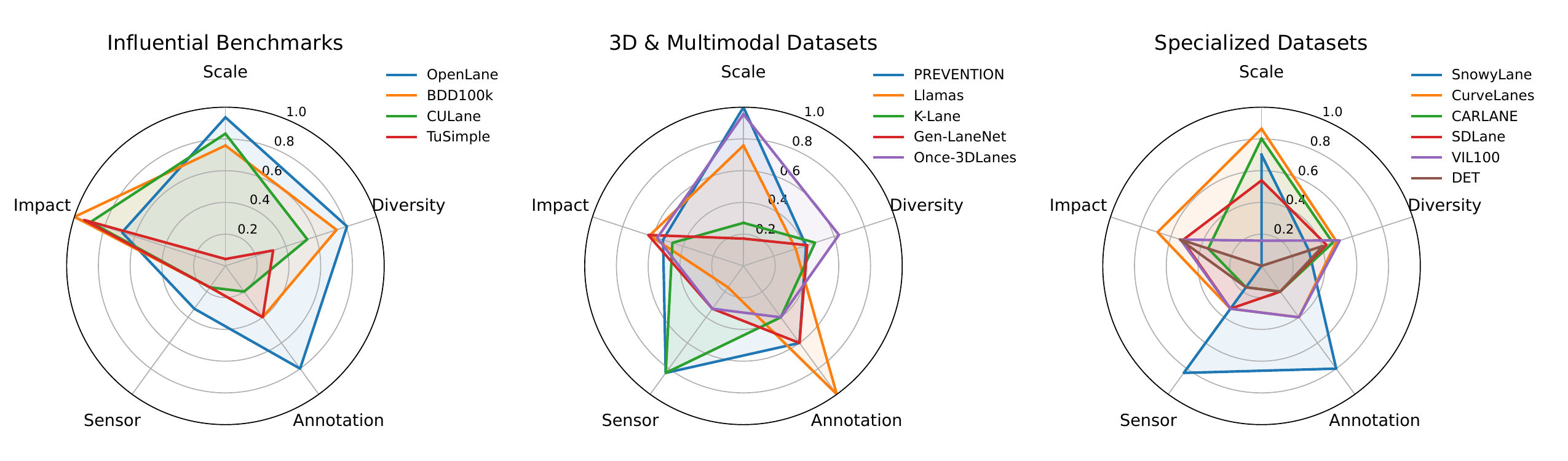}
    \caption{Scale, diversity, annotation, sensor and impact scores for influential benchmarks, 3D \& Multimodal datasets and specialized datasets.}
    \label{fig:radarplot}
    \vspace*{-5mm}
\end{figure*}

\begin{equation}
\label{eq:s_div}
S_{\text{div},d} = \frac{4 \cdot \frac{E_d}{E_{\max}}
+ 4 \cdot \frac{A_d}{A_{\max}}
+ 2 \cdot \frac{L_d}{L_{\max}}
+ I_{\text{real}}}{11}
\end{equation}

The environmental conditions and areas are weighted equally, as they mainly represent diversity. The number of lanes is for our work considered less relevant as it also correlated with the driving scenarios. real-world recordings receive a benefit as the data is more realistic as synthetic data.

Also the annotations are a KPI of datasets; hence $S_{\text{anno}}$ represents the normalized weighted annotation richness as shown in Eq.~\eqref{eq:anno}.

\begin{equation}
S_{\text{anno},d} =
\frac{
2 \cdot I_{2D,d}
+ 4 \cdot I_{3D,d}
+ 2 \cdot I_{\text{seg},d}
+ 2 \cdot I_{\text{cls},d}
}{10}
\label{eq:anno}
\end{equation}

with $I_{2D,d}$, $I_{3D,d}$, $I_{\text{seg},d}$ and 
and $I_{\text{cls},d}$ as binary indicator functions denoting the presence of 2D polylines, 3D polylines and segmentation masks,and at least five semantic classes, respectively. 3D labels are more informative and hence higher weighted.

Moreover, the sensor setup is considered as a KPI and a sensor score $S_{\text{sensor}}$ determined using Eq.~\eqref{eq:sensor}.
\begin{equation}
S_{\text{sensor},d} =
\frac{
I_{\text{RGB(S)},d}
+ 2\cdot I_{\text{RGB(H)},d}
+ 2\cdot I_{\text{LiDAR},d}
+ 1\cdot I_{\text{multi},d}
}{6}
\label{eq:sensor}
\end{equation}

with $I_{\text{RGB (SD)},d}$, $I_{\text{RGB (HD)},d}$, $I_{\text{LiDAR},d}$, and $I_{\text{multi},d}$ 
indicates the presence of RGB cameras (standard definition and high definition), LiDAR sensors, and multi-modal sensor setups (multi-camera or camera-LiDAR). Higher resolution cameras and advanced sensors as LiDARs are weighted higher due to enhanced sensing capabilities. Additionally, multi-sensor can further enhance the sensing capabilities and hence is considered as beneficial.

A final KPI is the research impact score $S_{\text{impact}}$, which shows the adoption and relevance of a dataset within the research community. To account for the fact that newer datasets have lower citation counts $C_d$ a logarithmic normalization as shown in Eq.~\eqref{eq:impact} is used.
\begin{equation}
S_{\text{impact},d} =
\frac{\log(C_d + 1)}
{\max_d (\log(C_d + 1))}.
\label{eq:impact}
\end{equation}

The final $\text{DQS}$ for dataset $d$ is defined as:
\begin{equation}
\label{eq:mdqs}
\begin{split}
    \text{DQS},d =&
0.30 \cdot S_{\text{div},d}
+ 0.20 \cdot S_{\text{anno},d}
+ 0.20 \cdot S_{\text{sensor},d}\\
+& 0.20 \cdot S_{\text{scale},d}
+ 0.10 \cdot S_{\text{impact},d}
\end{split}
\end{equation}
where each partial score $S_{\cdot,d} \in [0,1]$.

The weighting to determine $\text{DQS},d$ can vary depending on the goal the dataset should be used. For this work, the main goal is determine the suitability for robust lane detection under various conditions; hence, diversity $S_{\text{div},d}$ is assigned the highest weight (\SI{30}{\percent}) reflecting the dominant influence on generalization performance  and reliability under environmental scenarios and conditions such as adverse weather that significantly degrades perception. Annotations $S_{\text{div},d}$ and sensors $S_{\text{sensor},d}$ are considered as equally important as both contribute to the task complexity and the applicability of the dataset. Hence, both are assigned a weight of \SI{20}{\percent}. Equally important is the size of the dataset $S_{\text{scale},d}$, as a sufficient size is crucial to reliably train and evaluate lane detectors. Thus, also a weighting of \SI{20}{\percent} is used. As a final KPI the impact is incorporated as reflects community adoption and benchmarking relevance. However, this factor must be considered the least important and hence is only included with a weight of \SI{10}{percent}. This weighting scheme balances dataset size, complexity, and practical utility, yielding a robust and interpretable ranking regarding robust lane detection.
\section{RESULTS \& DISCUSSION}
\label{sec:results}
\begin{figure*}
    \includegraphics[width=\linewidth, trim={0cm 0cm 0cm 0cm}]{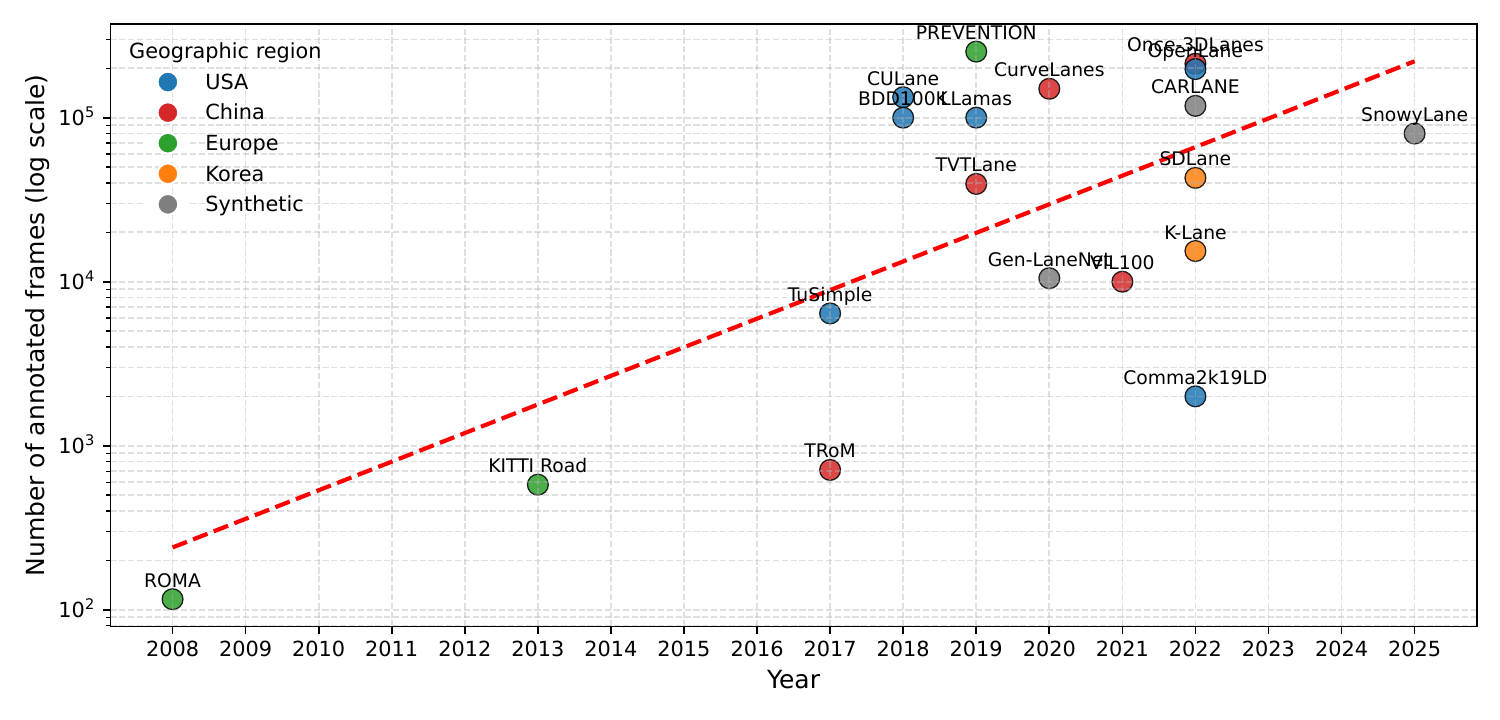}
    \caption{Evolution of dataset size including geographic region}
    \label{fig:sizes}
    \vspace*{-5mm}
\end{figure*}

The results of the evaluation are summarized in Tab.~\ref{tab:category_ranking} and visualized in Fig.~\ref{fig:radarplot} and Fig.~\ref{fig:sizes}. 
The ROMA, KITTI Road, TRoM, and Comma2k19LD datasets only feature between 116 and 2,000 frames, which is not a sufficient size for training state-of-the-art lane detection models. Hence, they can be considered as not suitable for robust lane detection. The TVTLane dataset provides a sufficient size of about 40,000 frames; however, their images are sampled to a really low resolution of 256$\times$128 px., which makes it unsuitable. Hence, these datasets are not further evaluated and discussed.

Regarding the influential benchmarks, OpenLane achieved the highest DQS with 0.733, achieving the best category and overall rank. This is mainly based on the rich annotations, the scale, and the higher diversity compared to the other datasets. BDD100K achieves a significantly lower DQS of 0.586, leading to overall rank 3. CULane achieved a DQS of 0.492, which leads to an overall rank 8. The difference to the other datasets within this category can be traced back to the low diversity in all categories. The TuSimple dataset achieved the lowest category result with only 0.310; this is mainly based on the limited diversity (only highway recordings at daytime) and the significantly lower number of frames compared to the other datasets.

For 3D \& Multimodal datasets, PREVENTION performed best with a DQS of 0.664. The dataset lacks adverse weather; however, it has a sufficient size and rich annotations and sensors. This leads to an overall rank 2 for PREVENTION. The ONCE-3DLanes with 0.585 shows a higher diversity in scenarios and environmental conditions; however, has less sensor modalities. Llamas and K-Lane achieve similar scores of 0.551 and 0.490, respectively. Both have a lower number of frames and less diversity, leading to the lower DQS compared to PREVENTION. Gen-LaneNet, with 0.411, performed worst within this category. The dataset has a low diversity, is only synthetic, and has a low size, which makes it less suitable for complex architectures.

\begin{table}[t]
\centering
\caption{Ranking of lane detection datasets using the proposed DQS framework with ranking within category (Cat. Rank) and overall (Overall).}
\label{tab:category_ranking}
\renewcommand{\arraystretch}{1.15}
\begin{tabular}{lllcc}
\toprule
\textbf{Category} & \textbf{Dataset} & \textbf{DQS}  & \textbf{Cat. Rank} & \textbf{Overall}\\
\midrule

\multirow{4}{*}{\textbf{\begin{tabular}[c]{@{}c@{}}Influential\\Benchmarks\end{tabular}}}
 &  OpenLane   & 0.733 & 1& 1\\
% &  Apolloscape& 0.641 & 2&3\\
 &  BDD100k    & 0.586 & 2&3\\
 &  CULane     & 0.492 & 3&8\\
 &  TuSimple   & 0.310 & 4&14\\
\midrule

\multirow{5}{*}{\textbf{\begin{tabular}[c]{@{}c@{}}3D \&\\Multimodal\\Datasets\end{tabular}}}
 &  PREVENTION     & 0.664 & 1& 2 \\
 &  Once-3DLanes   & 0.585 &2& 4\\
 &  Llamas         & 0.551 &3& 6\\
 &  K-Lane         & 0.490 &4& 9\\
 &  Gen-LaneNet    & 0.411 &5& 12\\
\midrule

\multirow{6}{*}{\textbf{\begin{tabular}[c]{@{}c@{}}Specialized\\ Datasets\end{tabular}}}
 & SnowyLane  & 0.561 & 1& 5\\
 & CurveLanes & 0.538 &2& 7\\
 & CARLANE    & 0.411 &3& 10\\
 & SDLane     & 0.396 &4& 12\\
 & VIL100     & 0.388 &5& 13\\
 & DET        & 0.250 &6& 15\\
 
\bottomrule
\end{tabular}
\vspace*{-5mm}
\end{table}

For the last category, the SnowyLane dataset achieved the best DQS score of 0.567, mainly benefiting from the rich annotations and sensor data, which leads to an overall rank of 5. Since the dataset is published in 2025, the impact is currently really low. The CurveLanes dataset achieved rank 7 with a DQS of 0.538, mostly based on the high number of frames and some diversity in scenarios and environmental conditions. CARLANE, VIL100 and SDLane all performed similarly, with a DQS ranging between 0.411 and 0.388. DET performed overall worst, with a DQS of only 0.150. This is based on the limited diversity in all categories, and mainly due to the low number of frames.

Overall, there is no perfect dataset that fits for all tasks, under all conditions. However, the OpenLane dataset comes really close. It is adopted in the community, which allows for a broad comparability, while having a high dataset size, which allows to even train and evaluate complex lane detection architectures. The dataset also shows a wide range of environments, and with up to 24 lanes and 15 classes of lane markings, the dataset outperforms all other datasets. Since rain and fog are included, the dataset is well suited for lane detection under adverse weather conditions. Additionally, the BDD100k is well suited due to the high availability of adverse weather conditions, and its wide adoption in research. The BDD100k dataset also features more tasks, such as object and traffic sign detection, which allows for a multi-use. CuLane and TuSimple are widely used; however, in 2025, they must be considered no longer suitable for lane detection applications due to the limited diversity regarding environments, conditions, and lanes. 
If LiDAR data is required, the PREVENTION dataset is the best choice. It has multi-camera and a LiDAR sensor, different environments, and some diversity in the lanes, and with 252,762 frames, a sufficient size also for complex architectures. Additionally, in contrast to most datasets, it is recorded in Europe extending the geographic diversity. Specialized datasets are also very helpful, as including everything in one dataset is very challenging, especially in real-world conditions. Hence, the Snowy Lane dataset is well suited for lane detection in winter conditions in rural environments. CurveLanes is also well suited for a more generalized application, as it features a high number of frames, but only in good weather conditions with 2D annotations. VIL100 provides fog data, but it is too small; hence, it should not be favored during the selection process. SDLane has the same goal as CurveLanes; however, significantly less frames and diversity. Hence, for curvy roads, CurveLanes should be chosen. CARLANE, due to the low complexity, can only be considered as a good choice for domain adaptation. However, for this purpose, a combination of other synthetic datasets with a higher ranked real-world dataset could lead to better results. For event-camera based approaches, the DET dataset is the only suitable dataset with reasonable sensor data, accurate labels, and a sufficient diversity regarding scenarios. However, the dataset is too small to train complex architectures.

\begin{table*}
\centering
\renewcommand{\arraystretch}{1.12}
\caption{Overview of Lane Detection Datasets and their key performance indicators. Citations are based on Google Scholar. For sensors it is one camera of type RGB and if not stated otherwise. Environmental conditions include day (D), night (N), clear weather (C), rain (R), fog (F), snow covering the road (SC) and snowfall (SF). (*) Citation count for TuSimple based on semantic scholar hits for "TuSimple Lane". Data source includes (Country Code) if real-world recording.}
\resizebox{\textwidth}{!}{%
\begin{tabular}{llllllllll}
\toprule
Dataset & Year & Citations & Frames & Sensor & Labels & Scenarios & Data Source & \#Lanes / \#Classes & Env. Conditions \\
%Caltech Lane~\cite{aly2008real}\begin{tabular}[c]{@{}l@{}}\ \\ \ \end{tabular} & 2008 &  1,224 & 640$\times$480 px. RGB& 2D Lines & Urban & real-world & $\leq$ 4 / - &\xmark & \xmark \\ \hline
%SLD~\cite{borkar2009robust}\begin{tabular}[c]{@{}l@{}}\ \\ \ \end{tabular} & 2009 &  2,021 & 720$\times$480 px. RGB& - & Urban, Highway & real-world & - / - &\xmark & \xmark \\ \hline
\midrule
\href{http://perso.lcpc.fr/tarel.jean-philippe/bdd/index.html}{ROMA}~\cite{veit2008evaluation}\begin{tabular}[c]{@{}l@{}}\ \\ \ \end{tabular} & 2008 &259&  116 & 1280$\times$1024 px. & 2D Segmentation & Rural & Real (FRA) & 2 / 2 & D, C \\ \hline
\href{https://www.cvlibs.net/datasets/kitti/eval_road.php}{KITTI Road}~\cite{Fritsch2013ITSC}\begin{tabular}[c]{@{}l@{}}\ \\ \ \end{tabular} & 2013 & 903&  579 & 2x 1242$\times$375 px.  & 2D Polygon & Urban & Real (GER) & 0-4 / 4 & D, C \\ \hline
%DIML~\cite{yoo2013gradient}\begin{tabular}[c]{@{}l@{}}\ \\ \ \end{tabular} & 2013 &  8,033 &  1280$\times$800 px. & Unknown & \begin{tabular}[c]{@{}l@{}}Urban, Rural, \\ Highway\end{tabular} & real-world & -/- &\checkmark & \xmark\\ \hline
\href{https://github.com/xllau/TRoM_annotation_v1.0}{TRoM}~\cite{liu2017benchmark}\begin{tabular}[c]{@{}l@{}}\ \\ \ \end{tabular} & 2017 & 404&  712 &  1280$\times$960 px. & 2D Segmentation & Highway & Real (CHN) & 1-4 / 4 & D, N, C, R, F \\ \hline
\href{https://www.kaggle.com/datasets/manideep1108/tusimple}{TuSimple}~\cite{tusimple}\begin{tabular}[c]{@{}l@{}}\ \\ \ \end{tabular} & 2017 & 2,920 (*) & 6,408 &  1280$\times$720 px. & 2D Polyline & Highway & Real (USA) & 2-4 / 7 &D, C \\ \hline
%VGPNet~\cite{Lee_2017_ICCV}\begin{tabular}[c]{@{}l@{}}\ \\ \ \end{tabular} & 2017 &  21,097 &  1288$\times$728 px. & 2D Polygons & Urban & real-world & - / 8 (17) &\checkmark & \xmark\\ \hline
%ELAS~\cite{berriel2017ego}\begin{tabular}[c]{@{}l@{}}\ \\ \ \end{tabular} & 2017 &  15k &  640$\times$480 px. & 2D Splines & Urban, Highway & real-world & - / 8 &\xmark & \xmark\\ \hline
\href{https://xingangpan.github.io/projects/CULane.html}{CULane}~\cite{pan2018SCNN}\begin{tabular}[c]{@{}l@{}}\ \\ \ \end{tabular} & 2018 & 1,493& 133,235 &  1640$\times$590 px. & 2D Polyline & \begin{tabular}[c]{@{}l@{}}Urban, Rural, \\ Highway\end{tabular} & Real (USA) & 0-3 / 4 & D, N, C \\ \hline
%Bai et al.~\cite{bai2018deep}\begin{tabular}[c]{@{}l@{}}\ \\ \ \end{tabular} & 2018 &  55,168 &  \begin{tabular}[c]{@{}l@{}} unknown \\ unknown LiDAR  \end{tabular}& 2D, 2D BEV& Urban, Highway & real-world &  \textbf{- / -} &\xmark & \xmark\\ \hline
\href{http://bdd-data.berkeley.edu/download.html}{BDD100k}~\cite{yu2020bdd100k}\begin{tabular}[c]{@{}l@{}}\ \\ \ \end{tabular} & 2018 & 3,570&  100,000 & 1280$\times$720 px. & 2D Polyline & \begin{tabular}[c]{@{}l@{}}Urban, Suburban, \\ Highway\end{tabular} & Real (USA) & 0-8 / 8   & D, N, C, R, F, SF \\ \hline
%\href{https://apolloscape.auto/index.html\#}{Apolloscape}~\cite{huang2018apolloscape}\begin{tabular}[c]{@{}l@{}}\ \\ \ \end{tabular} & 2018 & 1,446& 165,979 & 2x 1280$\times$720 px. & 2D \& 3D Segmentation & Urban, & Real (CHN) & - / 9  & D, N, C \\ \hline
%DSDLDE~\cite{lee2018robust}\begin{tabular}[c]{@{}l@{}}\ \\ \ \end{tabular} & 2018 &  33,323 & 1920$\times$1080 px. & 2D Polyline & Highway, Urban & real-world & - / - & \checkmark & \xmark \\ \hline
\href{https://prevention-dataset.uah.es/}{PREVENTION}~\cite{izquierdo2019prevention}\begin{tabular}[c]{@{}l@{}}\ \\ \ \end{tabular} & 2019 & 77 & 252,762 &  \begin{tabular}[c]{@{}l@{}} 2x 1920$\times$1200 px. \\ 32-layer LiDAR  \end{tabular}& 3D Polylines& Urban, Highway & Real (ESP)&  0-5 / 7 &D, C\\ \hline
\href{https://unsupervised-llamas.com/llamas/login/?next=/llamas/download}{Llamas}~\cite{llamas2019}\begin{tabular}[c]{@{}l@{}}\ \\ \ \end{tabular} & 2019 & 160&  100,042 & 1280$\times$717 px. & \begin{tabular}[c]{@{}l@{}} 2D \& 3D Polyline\\ 2D Segmentation \end{tabular}& Highway & Real (USA) & 2-8 / 6 & D, C \\ \hline
%\href{https://drive.google.com/drive/folders/1iO6EUira1_irMHnEFjma3vc8LfYKHmQ4}{Jiqing}~\cite{jiqing}\begin{tabular}[c]{@{}l@{}}\ \\ \ \end{tabular} & 2019 & -&  210,610 & 1x 1920$\times$1080 px. & 2D Polyline & Highway & Real & 2-6 / $\geq$ 2& D, C  \\ \hline
%Garnet et al. ~\cite{garnett20193d}\begin{tabular}[c]{@{}l@{}}\ \\ \ \end{tabular} & 2019 &  306k & 360$\times$480 px. & 3D Points & Highway & Synthetic &  - / - &\xmark & \xmark\\ \hline
\href{https://spritea.github.io/DET/}{DET}~\cite{det}\begin{tabular}[c]{@{}l@{}}\ \\ \ \end{tabular} & 2019 & 83 &  5,424 & 1280$\times$800 px. DVS& 2D Segmentation & Urban, Highway & Real (CHN) & 0-4 / 5& D, C \\ \hline
\href{https://github.com/qinnzou/Robust-Lane-Detection/blob/master/README.md}{TVTLane}~\cite{zou2019tvt}\begin{tabular}[c]{@{}l@{}}\ \\ \ \end{tabular} & 2019 & 494& 39,460 & 256$\times$128 px. & 2D Polyline & \begin{tabular}[c]{@{}l@{}}Urban, Rural, \\ Highway\end{tabular} & Real (CHN) & 2-4 / 7& D, C \\ \hline
\href{https://github.com/SoulmateB/CurveLanes}{CurveLanes}~\cite{xu2020curvelane}\begin{tabular}[c]{@{}l@{}}\ \\ \ \end{tabular} & 2020 & 278 &  150,000 &  2650$\times$1440 px. & 2D Polyline & Urban, Highway & Real (CHN) & 0-9 / 6 & D, N, C \\ \hline
\href{https://drive.google.com/file/d/1Kisxoj7mYl1YyA_4xBKTE8GGWiNZVain/view}{Gen-LaneNet}~\cite{guo2020gen}\begin{tabular}[c]{@{}l@{}}\ \\ \ \end{tabular} & 2020 & 173 & 10,500 &  1920$\times$1080 px. & 2D \& 3D Polyline & \begin{tabular}[c]{@{}l@{}}Urban, Suburban, \\ Highway\end{tabular} & Synthetic & 3-5 / 3 &D, C \\ \hline
%3DLanes~\cite{efrat2020semi}\begin{tabular}[c]{@{}l@{}}\ \\ \ \end{tabular} & 2020 &  327k / 299 k &  - & - & Urban, Highway & real-world & - / - &\xmark & \xmark\\ \hline
%TTLane~\cite{liang2020lane}\begin{tabular}[c]{@{}l@{}}\ \\ \ \end{tabular} & 2020 &  13,200 &  $\approx$2058$\times$1490 px. & 2D Bézier Curve & Urban & real-world & - / 5 &\checkmark & \xmark\\ \hline
%FusionLane~\cite{yin2020fusionlane}\begin{tabular}[c]{@{}l@{}}\ \\ \ \end{tabular} & 2020 &  437 &  \begin{tabular}[c]{@{}l@{}} 1242$\times$375 px. \\ 64-layer LiDAR  \end{tabular}& 2D BEV& Urban & real-world &  - / - &\xmark & \xmark\\ \hline
\href{https://github.com/yujun0-0/MMA-Net?tab=readme-ov-file}{VIL100}~\cite{zhang2021vil}\begin{tabular}[c]{@{}l@{}}\ \\ \ \end{tabular} & 2021 & 79 & 10,000 & $\approx$1920$\times$1080 px. & 2D Polyline & Urban, Highway & Real (CHN) & 1-5 / 10&D, N, C, F \\ \hline
\href{https://www.kaggle.com/datasets/tkm2261/comma2k19-ld}{Comma2k19 LD}~\cite{sato2022towards}\begin{tabular}[c]{@{}l@{}}\ \\ \ \end{tabular} & 2022 & 98 &  2,000 &  1920$\times$1080 px. & 2D Polyline & Highway & Real (USA) & 3-5 / 3 &D, C \\ \hline
 \href{https://once-3dlanes.github.io/3dlanes/}{Once-3DLanes}~\cite{yan2022once}\begin{tabular}[c]{@{}l@{}}\ \\ \ \end{tabular} & 2022 & 106 &  213,111 &  1920$\times$1080 px. & 3D Polyline & \begin{tabular}[c]{@{}l@{}}Urban, Suburban, \\ Highway\end{tabular} & Real (CHN) & 0-8 / 5 &D, N, C, R \\ \hline
 \href{https://github.com/kaist-avelab/k-lane}{K-Lane}~\cite{paek2022klane}\begin{tabular}[c]{@{}l@{}}\ \\ \ \end{tabular} & 2022 & 46 &  15,382 & \begin{tabular}[c]{@{}l@{}} 1920$\times$1200 px. \\ 64-layer LiDAR  \end{tabular}& 3D Polyline & Urban, Highway & Real (KOR) & 2-6 / 4& D, N, C  \\ \hline
 \href{https://github.com/OpenDriveLab/OpenLane/blob/main/data/README.md}{OpenLane}~\cite{chen2022persformer}\begin{tabular}[c]{@{}l@{}}\ \\ \ \end{tabular} & 2022 & 270 & 197,788 & 1920$\times$1080 px. & 2D \& 3D Polyline & \begin{tabular}[c]{@{}l@{}}Urban, Suburban, \\ Highway\end{tabular} & Real (USA) & 0-24 / 15 & D, N, C, R, F \\ \hline
%\href{https://github.com/Thinklab-SJTU/OpenDenseLane}{OpenDenseLane}~\cite{chen2022opendenselane}\begin{tabular}[c]{@{}l@{}}\ \\ \ \end{tabular} & 2022 &  57,227 &  \begin{tabular}[c]{@{}l@{}} unknown \\3x 128-layer LiDAR  \end{tabular}& 2D BEV \& 3D Polyline & Urban, Highway & real-world & 0-8 / $\geq$ 6 &\xmark \\ \hline
\href{https://www.kaggle.com/datasets/carlanebenchmark/carlane-benchmark}{CARLANE}~\cite{NEURIPS2022_19a26064}\begin{tabular}[c]{@{}l@{}}\ \\ \ \end{tabular} & 2022 & 17 & 118,000& 1280$\times$720 px. & 2D Polyline & Urban, Highway & Synthetic \& Real & 1-4 / 3  &D, N, C, R, F \\ \hline
\href{https://www.42dot.ai/akit/dataset/sdlane/download}{SDLane}~\cite{jin2022eigenlanes}\begin{tabular}[c]{@{}l@{}}\ \\ \ \end{tabular} & 2022 & 71 & 42,949 &  1920$\times$1208 px. & 2D Polyline & Urban, Highway & Real (KOR) & 0-7 / 4 &D, C \\ \hline
%Simulanes~\cite{hu2022sim}\begin{tabular}[c]{@{}l@{}}\ \\ \ \end{tabular} & 2022 &  16,344 &  1280$\times$720 px. & 2D Polylines & \begin{tabular}[c]{@{}l@{}}Urban, Suburban, \\ Rural, Highway\end{tabular} & Synthetic & 2-10 / 15 &\xmark & (\xmark) \\ \hline
%LanEvil~\cite{zhang2024lanevil}\begin{tabular}[c]{@{}l@{}}\ \\ \ \end{tabular} & 2024 &  90,292 &  1280$\times$720 px. & 2D Polylines & Urban, Highway & Synthetic &  1-5 / 9 &\checkmark & \xmark\\ \hline
\href{https://ekut-es.github.io/snowy-lane/}{SnowyLane}~\cite{gamerdinger2025snowylane}\begin{tabular}[c]{@{}l@{}}\ \\ \ \end{tabular} & 2025 & 0 &80,000 &  \begin{tabular}[c]{@{}l@{}} 1920$\times$1080 px. \\32-layer LiDAR \end{tabular}&\begin{tabular}[c]{@{}l@{}} 2D \& 3D Polyline \\
2D Segmentation\end{tabular}& Rural & Synthetic & 2 / 3 & D, C, SC, SF \\

%-----------------------------------------------------------------------------------------------------------
\bottomrule
\end{tabular}
}
\vspace*{-4mm}
\label{tab:overview}
\end{table*}

%%%%%%%%%%%%%%%%%%%%%%%%%%%%%%%%%%%%%%%%%%%%%%%%%%%%%%%%%%%%%%%%%%%%%%%%%%%%%%%%
\section{CONCLUSION \& OUTLOOK}
\label{sec:conclusion}
In this paper, we present a comprehensive review of lane detection datasets. We included a total of 20 published and publicly available datasets, present in chronological order. We provide key performance indicators such as dataset size, sensor modalities, annotation formats, scenario diversity, and environmental conditions. We have proposed an evaluation metric to provide deeper insight into the properties of the available datasets, and we have discussed these datasets extensively. Our evaluation and discussion revealed that there are already a wide variety of suitable datasets. We then propose a novel multidimensional dataset quality score to extensively compare the presented datasets and rank these datasets.
In addition to evaluating the datasets, we identified different trends. To achieve market readiness and train increasingly complex deep learning models, more training data is required. This has led to a significant increase in dataset size, from a few hundred frames for early datasets to over 250,000 frames for more recent ones (see Fig.~\ref{fig:sizes}). While early datasets only provided low-resolution images, newer datasets always provide high-resolution images and sometimes LiDAR data. Geographic bias may occur when selecting datasets, as most datasets (12 out of 17 real-world datasets) are recorded in the United States or China. Combining datasets can be a useful method of achieving better generalization. Incorporating synthetic datasets can also increase robustness, since simulators can generate controllable corner cases and weather conditions, as well as providing 3D information, which is difficult to generate in real-world environments. Due to these advantages, the proportion of synthetic datasets has increased in recent years.
Further research should overcome the identified limitations of existing datasets by providing sufficient size and highly diverse environments. A promising approach would be to augment existing datasets with adverse weather conditions using generative artificial intelligence, since adverse weather conditions are difficult to capture in various regions of the world and this would allow the use of already available datasets to be enhanced. More advanced sensing technologies, such as LiDAR and event cameras are also promising for improving lane detection, but are less common across datasets. Additionally, extending the ground truth with map data and vehicle state information enables the incorporation of enhanced evaluation metrics, further improving the safety and robustness of lane detection.

\bibliographystyle{IEEEtran} % use IEEEtran.bst style
\bibliography{literature.bib}

\end{document}